\documentclass{bmvc2k}

\usepackage{makecell}
\usepackage{multirow}
\title{Universal Perturbation Attack \\ on Differentiable No-Reference \\Image- and Video-Quality Metrics}

\addauthor{Ekaterina Shumitskaya}{ekaterina.shumitskaya@graphics.cs.msu.ru}{1}
\addauthor{Anastasia Antsiferova}{aantsiferova@graphics.cs.msu.ru}{1, 2}
\addauthor{Dmitriy Vatolin}{dmitriy@graphics.cs.msu.ru}{1, 2}

\addinstitution{
 Lomonosov Moscow \\ State University\\
 Moscow, RU
}

\addinstitution{
 MSU Institute for Artificial \\ Intelligence\\
 Moscow, RU
}

\runninghead{Shumitskaya, Antsiferova, Vatolin}{Universal Attack on NR Metrics}

\def\etal{\emph{et al}\bmvaOneDot}

\begin{document}

\maketitle

\begin{abstract}
Universal adversarial perturbation (UAP) attacks are widely used to analyze image classifiers that employ convolutional neural networks. In this paper, we make the first attempt in attacking differentiable no-reference image- and video-quality metrics through UAPs. The goal of attacks on quality metric is to increase the quality score of an output image, when visual quality does not improve after the attack. The development of new attacks plays an important role in vulnerability analysis of quality metrics. When developers of image- and video-algorithms can boost metric scores through preprocessing, objective algorithm comparisons are no longer fair. Inspired by the idea of UAPs for classifiers, we trained UAPs for seven no-reference image- and video-quality metrics (PaQ-2-PiQ, Linearity, VSFA, MDTVSFA, KonCept512, Nima and SPAQ) to increase the respective scores. We treated the UAP as network weights and applied the deep-learning training techniques. We then applied trained UAPs to FullHD video frames before compression and proposed a method for comparing metrics stability based on RD curves to identify metrics that are the most resistant to UAP attack. The existence of successful UAP appears to diminish metric’s ability to provide reliable scores. We recommend the proposed method as an additional verification of metric reliability to complement traditional subjective tests and benchmarks.
\end{abstract}

%-------------------------------------------------------------------------
\section{Introduction}
\label{sec:intro}
Video-quality assessment has always been an important task for processing and transmitting video over the Internet. But quality assessment using subjective evaluations is expensive and time-consuming hence the creation of new quality assessment algorithms. Objective quality metrics are common in the development and comparison of image- and video-processing algorithms. Depending on the original video’s availability, these metrics fall into three categories: full-reference (FR), reduced-reference (RR) and no-reference (NR). NR metrics are becoming more popular owing to greater applicability than FR and RR metrics, since in many real-life cases the reference video is unavailable. NR metrics are widely used for tasks such as collecting content-quality statistics for real-time communication and streaming, in addition to optimizing image- and video-processing parameters \cite{shahid2014no}. State-of-the-art NR metrics commonly use neural-network-based approaches. They often deliver more accuracy than traditional approaches, but they are also more vulnerable to attacks that increase their output scores. Most video-processing algorithms, for example encoding, deblurring, denoising and super-resolution can optimize images or video frames in accordance with the given NR metric and increase the score.

A traditional way to evaluate quality-metric performance is subjective testing that is measuring the correlation of metric scores with subjective scores for some dataset. But just checking the correlation is insufficient for NR metrics. Estimating other parameters is important: for example, the ability to provide reliable scores that are immune to increases through detached processing. Metric-score reliability is crucial to comparing algorithms, but NR metrics are vulnerable to processing that increases the output score because for each image or video, the direction of increasing scores depends only on this image or video (without reference image or video).

\begin{figure*}[htb]
\begin{center}
\includegraphics[width=9.5cm]{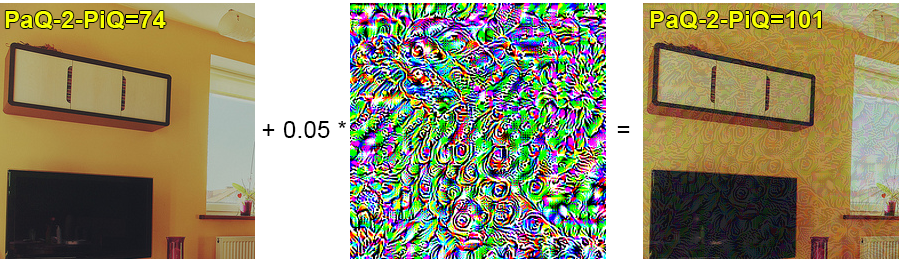}
\end{center}
   \caption{Universal perturbation attack on the no-reference quality metric PaQ-2-PiQ.}
\label{fig:upa}
\end{figure*}

In this paper, we analyze the vulnerability of new NR quality metrics to score increases by applying universal perturbation attacks. Our work defines universal perturbation as a fixed perturbation trained for a given NR quality metric to increase its scores when applied to images or videos (frame by frame). Figure \ref{fig:upa} illustrates a universal perturbation attack on the NR metric PaQ-2-PiQ \cite{Ying_2020_CVPR}. Moosavi-Dezfooli et al. \cite{Moosavi-Dezfooli_2017_CVPR} previously introduced universal adversarial perturbation as a fixed perturbation that can fool a convolutional-neural-network-based image classifier. We employed a universal perturbation attack on seven differentiable NR metrics, estimating their resistance to this attack using our proposed stability score. Stability-evaluation method works for any type of universal attacks that is when the same transformation is applied to all images or videos. It can also serve as complementary performance testing for quality metrics. We make our code publicly available at: \url{https://github.com/katiashh/UAP_Attack_on_Quality_Metrics}.

%Our main contributions in this paper are the following:
%\begin{itemize}
%\item A new method of attack on differentiable no-reference image- and video-quality metrics through trainable universal perturbation.
%\item A new method for assessing the stability of image- and video- quality metrics that complements traditional subjective tests.
%\end{itemize}

\section{Related Work}
The most popular way to evaluate the performance of an image- or video-quality metric is to conduct a subjective test. For such tests, the researchers create a special large dataset with various distortions and measure the correlation between quality scores and subjective scores for that dataset. Owing to the complexity and high cost of subjective evaluations, researchers have proposed several ways to reduce dataset size or fully automate the process of measuring quality metric performance.

Ciaramello and Reibman \cite{ciaramello2011supplemental} introduced a method based on generation of image or video pairs with the intent to cause misclassification errors by a quality metric; they then conducted small-scale pairwise subjective testing. They also presented a systematic computational black-box test for image-quality metrics to quantify many of their vulnerabilities \cite{ciaramello2011systematic}. They obtained metric’s performance information for a constrained set of degraded images and then identified metric-score inconsistencies on the basis of these results. Liu and Reibman \cite{liu2016software} introduced software, called “STIQE” that automatically explores an image-quality metric’s performance. It allows users to execute a series of tests and then generate reports to determine how well the metric performs. Testing consists of applying to images several distortions of varying degrees and checking whether the metric score rises monotonically as the applied distortion increases. Wang and Simoncelli \cite{wang2008maximum} proposed a method for comparing computational models of perceptual quantities called maximum differentiation competition (MADC). It can compare two image- and video-quality metrics by constructing a pair of examples that maximize or minimize the score of one metric while keeping the other fixed. It then repeats this procedure, but reverses the roles of the two quality metrics. To identify weaknesses in a quality metrics, subjective testing on pairs of such constructed examples is necessary. If the scores for one metric correlate better with human judgment than the scores for another metric on these pairs, the first metric is considered as more stable. Kede Ma \etal \cite{Ma_2016_CVPR} extended the idea of MADC by proposing group maximum differentiation competition (gMADC). This technique employs a group of quality models (“attackers”) to automatically generate examples that maximize or minimize the attackers scores, while keeping scores of another resistant model fixed (“defender”). For example, Zhang \etal \cite{zhang2018blind} demonstrated the stability of their proposed NR quality metric using gMADC. Kettunen \etal \cite{kettunen2019lpips} pursued a similar idea and showed that the LPIPS learned perceptual similarity metric is susceptible to attack by constructing image pairs that humans perceive as dissimilar but LPIPS considers close, and vice versa by solving optimization problems.

Existing quality metric evaluation methods either do not check metric resistance to score increases or cannot operate in real time with real algorithms that process and compress images or videos. At first, we intended to test quality metric stability using MADC: we maximized metrics scores with constraints on MSE. However due to a high computational complexity of MADC, we were looking for a way to accelerate it without significant changes of metric scores before and after the attack. The speed of attack is very important. Slow attacks have less practical meaning since they are unlikely to be injected into video compression and processing algorithms. Due to high-speed importance, we decided to use the input-agnostic universal perturbation attack instead of the input-dependent direct gradient attack for stability testing. To apply a universal attack it is enough to simply add perturbation to the video frames. 

%Existing quality metric evaluation methods either do not check metric resistance to score increases or cannot operate in real time with real algorithms that process and compress images or videos. Thus, it is necessary to check for the possibility of a fast attack that can fool quality metrics and to estimate resistance to this type of attack. 
%At first, we intended to test quality metric stability using an attack based on MAD competition: maximize the metric scores with MSE constraints. But MAD competition has high computational complexity, and we were looking for a way to accelerate it without significantly changing the difference in metric scores before and after the attack. Thus, for each metric we trained a convolutional neural network and a universal perturbation to increase the scores for images or video frames. The universal perturbation attack turned out to be faster yet yielded nearly the same metric score differences. Therefore, we selected it for stability testing. 

\section{Proposed Methods}
\label{sec:proposed_method}
\begin{figure*}[htb]
\begin{center}
\includegraphics[width=10.5cm]{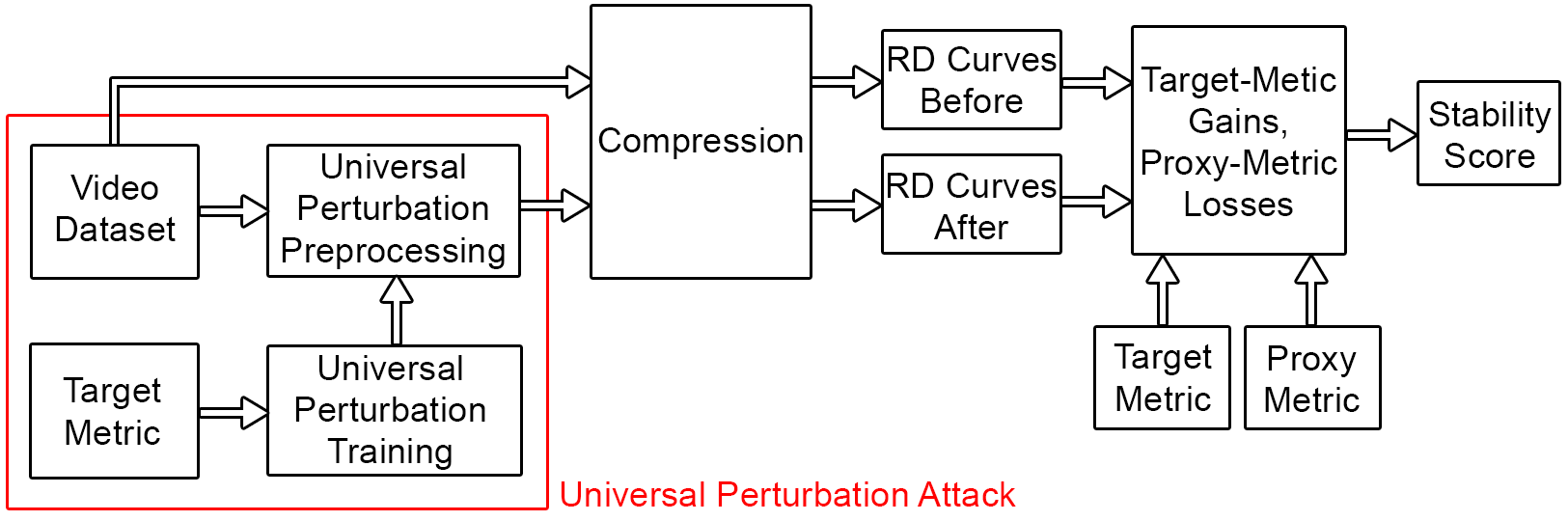}
\end{center}
   \caption{Scheme of proposed method for assessing target metric stability.}
\label{fig:method_scheme}
\end{figure*}

Our method for assessing NR quality metric stability consists of two components: the attack method and the stability score calculation. In this paper we use the terms {\em target metric} and {\em proxy metric}. A target metric is a metric whose stability requires scrutiny. A proxy metric is a metric for estimating visual quality degradation. We chose PSNR as our proxy metric, because it is the simplest quality metric and is difficult to cheat through processing. Figure \ref{fig:method_scheme} presents the general scheme of our method. For each target metric we trained a universal perturbation that can increase the score when applied to the original images or videos frames. We then applied the perturbation to frames in a set of uncompressed videos, compressed each video and drew rate-distortion (RD) curves for target and proxy metrics. Having RD curves with and without perturbation preprocessing allowed us to estimate the difference between target and proxy metric scores and calculate the stability score. 
%Section \ref{sec:results} provides a comparison of all seven tested metrics by stability score.

\subsection{Universal perturbation attack}
\label{sec:attack_method}
Video compression and processing takes place at high resolution. However, training of FullHD (1920x1080) universal perturbation can be replaced by training low-resolution perturbation (256x256) and then constructing high-resolution perturbation by repeating low-resolution perturbation. Quality metrics are also often trained on low-resolution images  \cite{li2020norm}, \cite{Fang_2020_CVPR}, \cite{talebi2018nima}, because training on low-resolution allows the use of batch training and leads to better convergence.

We trained a 256×256 universal perturbation on 10,000 images of the same size from the COCO dataset \cite{lin2014microsoft}. During the training, we cropped the perturbation’s values to the range -0.1 to 0.1 (the image-pixel values range from 0 to 1). Let {\em batch} be a batch of images for training and let {\em p} be a trainable universal perturbation; p is a real tensor whose values are adjusted with respect to increasing scores from the target NR metric. We can write the loss function as equation \ref{eq:1}, where {\em a} is a normalization factor. We also tried loss with the RMSE component, but this led to an uneven distribution of the perturbation pattern (depends on training data). We employed the Adam optimizer \cite{kingma2014adam} with a learning rate of 0.001 and trained our universal perturbation over five epochs. The batch size was eight. We then obtained universal perturbations with different amplitude levels by multiplying values of trained perturbation by a scaling factor.

\begin{equation} \label{eq:1}
loss = 1 - \frac{target \ metric(batch + p)}{a}
\end{equation}

We compared our attack method with a gradient attack, performed using MADC \cite{wang2008maximum} method on the NR metric PaQ-2-PiQ.  Universal perturbation is an approximation for perturbation produced by MADC, but is much faster to execute. The first row in Table \ref{table:tab1} presents results of applying our PaQ-2-PiQ universal perturbation to 200 other images of size 256×256 from the COCO dataset with eight amplitude levels (0.02, 0.04, 0.06, 0.08, 0.1, 0.2, 0.4 and 0.6). The second row contains results of applying to images a MADC attack with the same amplitude levels, where PaQ-2-PiQ increases and MSE remains below the amplitude level. For each image we conducted 1,000 gradient-descent steps. The MADC attack further increases the scores, but it also takes much longer, as it requires several backpropagation steps. In our experiments, on average, the universal perturbation attack yields 70\% of the metric score boost that the MADC attack achieved, but in just 0.1\% of the time. The high speed of UAP attack makes it applicable in real cases, like video compression or video processing. 

\begin{table}[htb]
\begin{center}
\begin{tabular}{|c | c c c c c c c c| c c |}
\hline
 & \multicolumn{8}{c}{\textbf{PaQ-2-PiQ difference}} & \multicolumn{2}{|c|}{}\\

\cline{2-11}
\multicolumn{1}{|c|}{\makecell{\textbf{Amplitude} \\ \textbf{level}}}
&
\makecell{0.02} & 0.04 & 0.06 & 0.08 & 0.1 & 0.2 & 0.4 & 0.6 & \makecell{\textbf{CPU} \\ \textbf{time}} & \makecell{\textbf{GPU} \\ \textbf{time}}\\
\hline 
\makecell{Universal\\ perturbation \\ attack mean} & 4.2 & 17.1 & 27.7 & 35.8 & 42.3
& 60.9 & 74.2 & 79.1 & 2 sec & 2 sec \\ 
\hline
\makecell{MADC \cite{wang2008maximum} \\ attack mean} & 28.2 & 39.9 & 47.3 & 52.7 & 56.8
& 69.1 & 76.8 & 77.8 & \makecell{1200 \\ min} & 32 min \\ 
\hline
\end{tabular}
\caption{PaQ-2-PiQ average differences before and after applying universal perturbation and MADC attacks to 200 tested images. The metric scores for the original images are subtracted from the metric scores after the attack.}
\label{table:tab1}
\end{center}
\end{table}

\begin{figure*}[htb]
\begin{center}
\includegraphics[width=11cm]{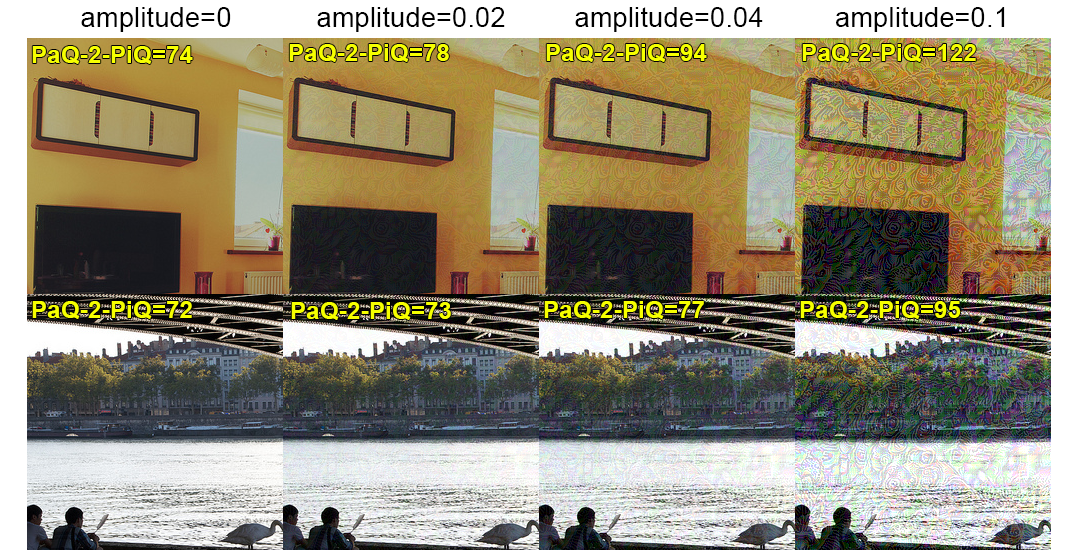}
\end{center}
   \caption{Different PaQ-2-PiQ universal perturbation amplitudes applied to images. The higher the amplitude, the greater the PaQ-2-PiQ score, but visual distortion also becomes more apparent.}
\label{fig:p2p_examples}
\end{figure*}

On the basis of those results we conclude that for a convolutional NR quality metric, there may be a general gradient whose direction corresponds to increasing metric scores. And universal perturbation learns this general direction. Figure \ref{fig:p2p_examples} presents images before and after addition of the PaQ-2-PiQ universal perturbation with three amplitude levels.

\begin{figure*}[htb]
\begin{center}
\includegraphics[width=11cm]{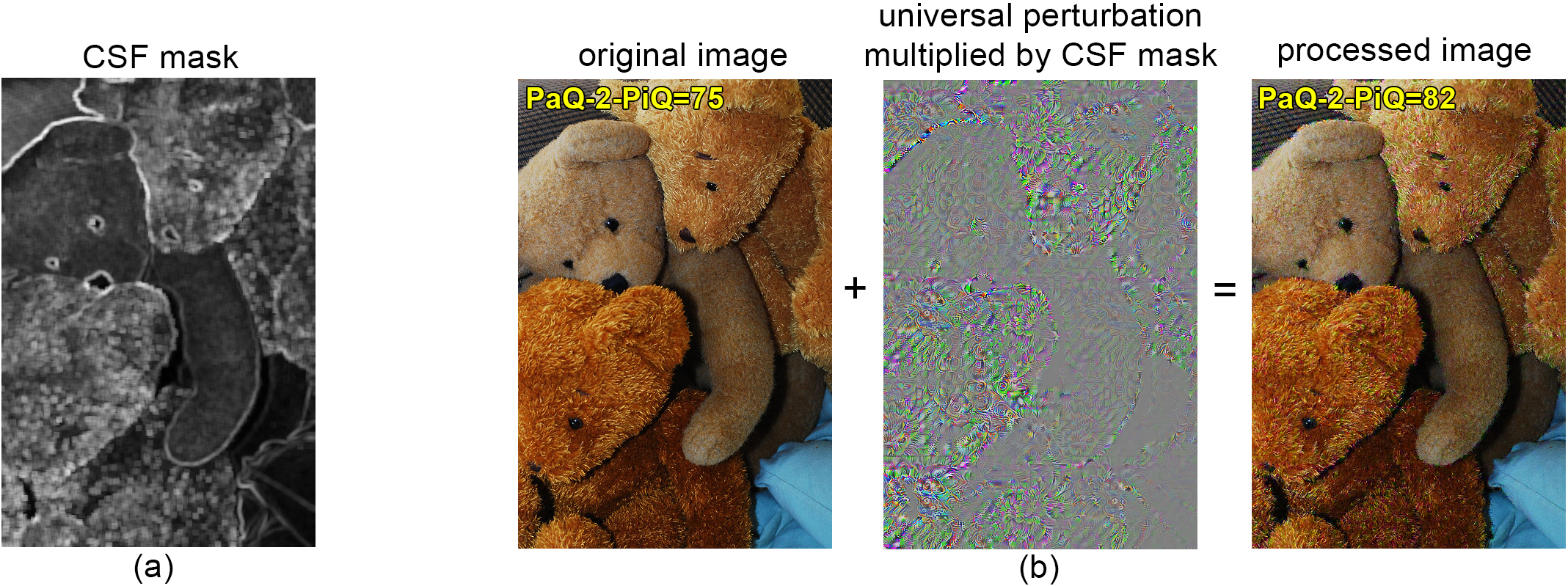}
\end{center}
   \caption{CSF mask (a) and proportional application of PaQ-2-PiQ universal perturbation to the CSF mask (b).}
\label{fig:csf_example}
\end{figure*}

We performed a subjective analysis of images and videos produced by universal perturbation attacks. The distortions are uniform and noticeable enough at high amplitude levels. There are ways, however, to reduce the visibility of a uniform universal perturbation. For example, we applied the universal perturbation in proportion to the contrast because people are less likely to notice distortion in high-contrast areas. We calculated the contrast sensitivity function (CSF) mask (Figure \ref{fig:csf_example} (a)), multiplied the PaQ-2-PiQ universal perturbation by this mask and added the result to the original image (Figure \ref{fig:csf_example} (b)). As the figure shows, such a transformation can still increase the PaQ-2-PiQ scores, but at the same time it is almost invisible. 
%The visibility of the universal perturbation depends on the contrast of the image or video frame. For example, encoder can apply perturbation with different amplitude depending on the contrast.

\subsubsection{Analysis of training on different data}
To establish that the trained universal perturbations are not random outliers, and that they are easily trainable on other data and can achieve nearly the same results, we conducted an ablation study. We examined the influence of the following factors: training multiple times on the same data, training with different data, training with different dataset sizes and training with different numbers of epochs. The supplemental materials provide graphs of our results.

\subsection{Stability score}
\label{sec:stability_score}
%We applied a trained universal perturbation to 20 FullHD raw videos from the Xiph.org dataset \cite{xiph} at four amplitude levels (0.02, 0.04, 0.06 and 0.08). 
%Figure \ref{fig:rd_p2p} (a) presents the resulting PaQ-2-PiQ RD curves for four videos, and Figure \ref{fig:rd_p2p} (b) presents the PSNR RD curves.
%We then compressed each video (including the originals) using the H.264 codec. The higher the RD curve, the more the attack increased the metric score.
Testing the effectiveness of the universal perturbation attack on videos requires considering compression, because most videos are compressed for transmission and storage. To apply a trained universal perturbation (see Section \ref{sec:attack_method}) to high-resolution video, we composed another high-resolution universal perturbation of the necessary size from a low-resolution 256x256 trained universal perturbation square. We applied universal perturbation with $K$ amplitude levels to dataset that contains $N$ videos, compressed each video (including the originals) and calculated the RD curves with and without preprocessing for target and proxy metrics. The higher target-metric RD curve with preprocessing, the more the attack increased the metric score. But different metrics exhibit different score ranges. To compare score differences before and after the attack, we therefore introduce the concepts of target-metric gain and proxy-metric loss and calculate the stability score.

At first, for each of the $M$ target metrics we employed the following procedure for target-metric gain versus proxy-metric loss dependence construction:

Step 1. Calculate the maximum and minimum target-metric scores on the RD curves for $N$ videos ($K+1$ curves for each video – $(K+1)*N$ curves in total). The result is the range of target-metric scores for our dataset. 

Step 2. Do min/max normalization of all target-metric RD curves to a range of 0–1 (along the y-axis) by subtracting the minimum and then dividing by difference between the maximum and the minimum. Regardless of the metric, the result is RD curves where the target-metric scores range from 0 to 1. We interpret the normalized RD curves as follows: the minimum metric score on our dataset corresponds to 0 and the maximum value corresponds to 1. 

Step 3. As in steps 1 and 2, do min/max normalization of all proxy-metric RD curves. We calculate the minimum and maximum value of the proxy metric using $M*N*(K+1)$ RD curves.

Step 4. Define the gain for a specific target-metric RD curve with preprocessing and a certain amplitude level as the difference between the area under that RD curve and the area under the target-metric RD curve without preprocessing. We calculated areas for the region where both RD curves are determined, replacing the x-axis by the segment [0, 1]. The target-metric gain shows the percent increase of the target-metric range after preprocessing.

Step 5. As in step 4, define the loss for a specific proxy-metric RD curve with preprocessing and a certain amplitude level as the difference between the area under that RD curve and the area under the proxy-metric RD curve without preprocessing.

Step 6. Average the target-metric gains and proxy-metric losses over $N$ videos for each amplitude level.

Step 7. Draw target-metric gain versus proxy-metric loss dependence, which consists of $K$ points corresponding to $K$ amplitude levels.

\begin{figure*}[htb]
\begin{center}
\includegraphics[width=12cm]{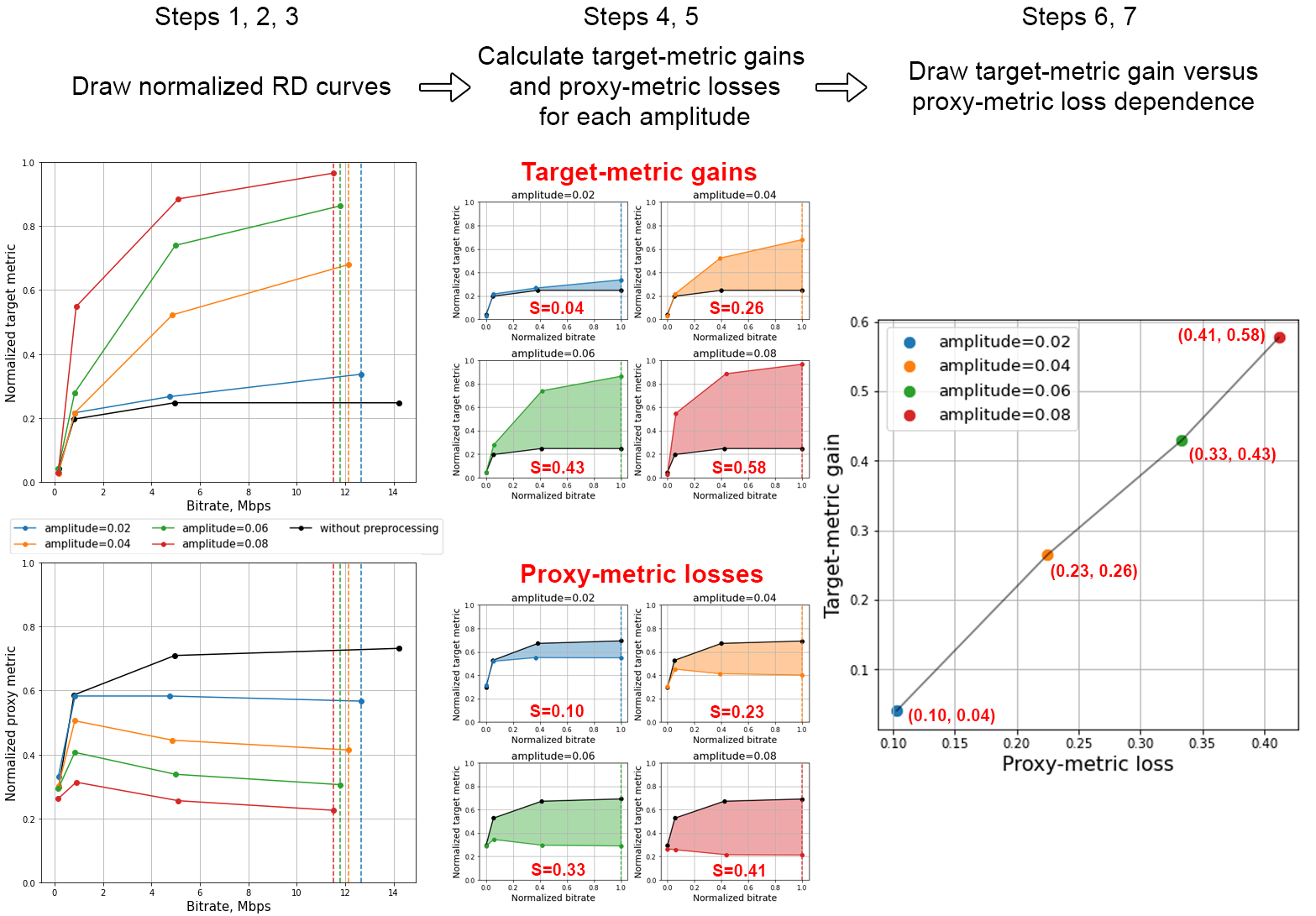}
\end{center}
   \caption{Calculation of target-metric gain and proxy-metric loss using normalized RD curves for a video at four amplitude levels. The rightmost chart shows the target-metric versus proxy-metric loss dependence for these amplitude levels.}
\label{fig:tmg_and_pml}
\end{figure*}

Figure \ref{fig:tmg_and_pml} presents an example calculation of target-metric gain and proxy-metric loss from normalized RD curves for four amplitude levels on one video. It also shows the target-metric gain versus proxy-metric loss dependence, which we constructed for these amplitudes. Given this dependence for each metric, we can compare metrics with each other. The dependence of a stable no-reference metric has a zero or negative slope, meaning application of a universal perturbation should either leave the metric score unchanged or reduce it. For vulnerable metrics, the dependence has a positive slope, since trained universal perturbations boost metric scores in these cases. 

\begin{figure*}[htb]
\begin{center}
\includegraphics[width=9cm]{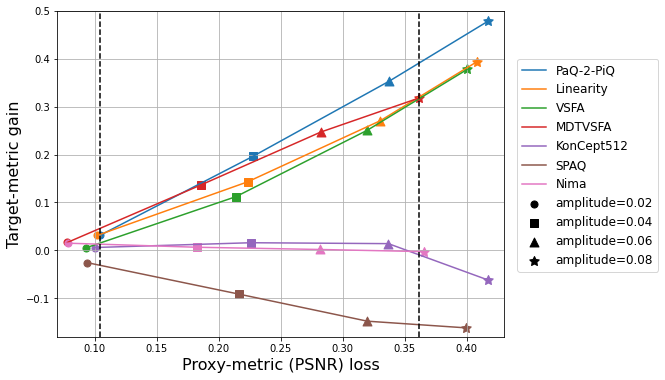}
\end{center}
   \caption{Target-metric gain versus proxy-metric loss dependencies for all tested NR metrics. Dotted lines highlight the region where dependencies are defined for all metrics. The less target-metric gain and the higher PSNR loss are, the more stable is the metric.}
\label{fig:res_dep}
\end{figure*}

We construct target-metric gain versus proxy-metric loss dependencies for 7 NR metrics (PaQ-2-PiQ \cite{Ying_2020_CVPR}, Linearity \cite{li2020norm}, VSFA \cite{li2019quality}, MDTVSFA \cite{li2021unified}, KonCept512 \cite{hosu2020koniq}, Nima \cite{talebi2018nima} and SPAQ \cite{Fang_2020_CVPR}) (see Figure \ref{fig:res_dep}). We used 20 FullHD raw videos from the Xiph.org dataset \cite{xiph}, four amplitude levels (0.02, 0.04, 0.06 and 0.08) and compressed videos using the H.264 codec preset medium at four bitrates (200 kbps, 1 Mbps, 5 Mbps and 12 Mbps).

To characterize the resistance of each metric to attack by a single number, we introduce a stability score as the area under the target-metric gain versus proxy-metric loss dependence, with the opposite sign, multiplied by 100. For each metric we calculate the area along the proxy-metric-axis where dependencies for all tested metrics are determined. A negative stability score corresponds to vulnerable metrics, while a positive stability score corresponds to stable metrics. Greater stability scores are better because the metric correctly responds to the addition of universal perturbation, which spoils original images and videos. Section \ref{sec:results} presents rating of tested NR metrics by proposed stability score.

%\begin{figure*}[htb]
%\begin{center}
%\includegraphics[width=11cm]{images/new_rd_curves_p2p_v2.png}
%\end{center}
%   \caption{PaQ-2-PiQ RD curves (a) and PSNR RD curves (b) for four videos.}
%\label{fig:rd_p2p}
%\end{figure*}
%Section \ref{sec:results} presents stability scores as well as target-metric-gain versus proxy-metric-loss dependencies for seven NR quality metrics. 

\section{Results}
\label{sec:results}
\begin{figure*}[htb]
\begin{center}
\includegraphics[width=12.5cm]{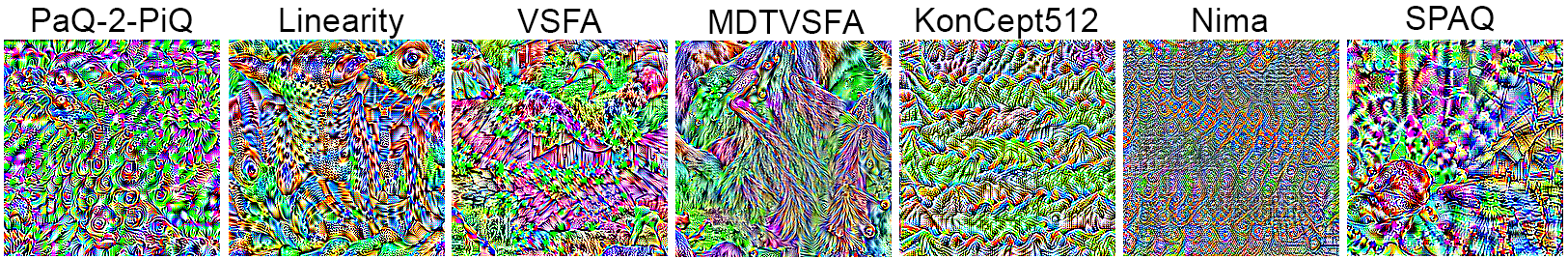}
\end{center}
   \caption{Trained universal perturbations for all seven tested metrics.}
\label{fig:up_vis}
\end{figure*}

We trained universal perturbations for seven differentiable no-reference quality metrics (PaQ-2-PiQ \cite{Ying_2020_CVPR}, Linearity \cite{li2020norm}, VSFA \cite{li2019quality}, MDTVSFA \cite{li2021unified}, KonCept512 \cite{hosu2020koniq}, Nima \cite{talebi2018nima} and SPAQ \cite{Fang_2020_CVPR}) on 10,000 images from the COCO dataset \cite{lin2014microsoft}. Figure \ref{fig:up_vis} shows visualizations of the trained universal perturbations.

\begin{figure*}[htb]
\begin{center}
\includegraphics[width=10.5cm]{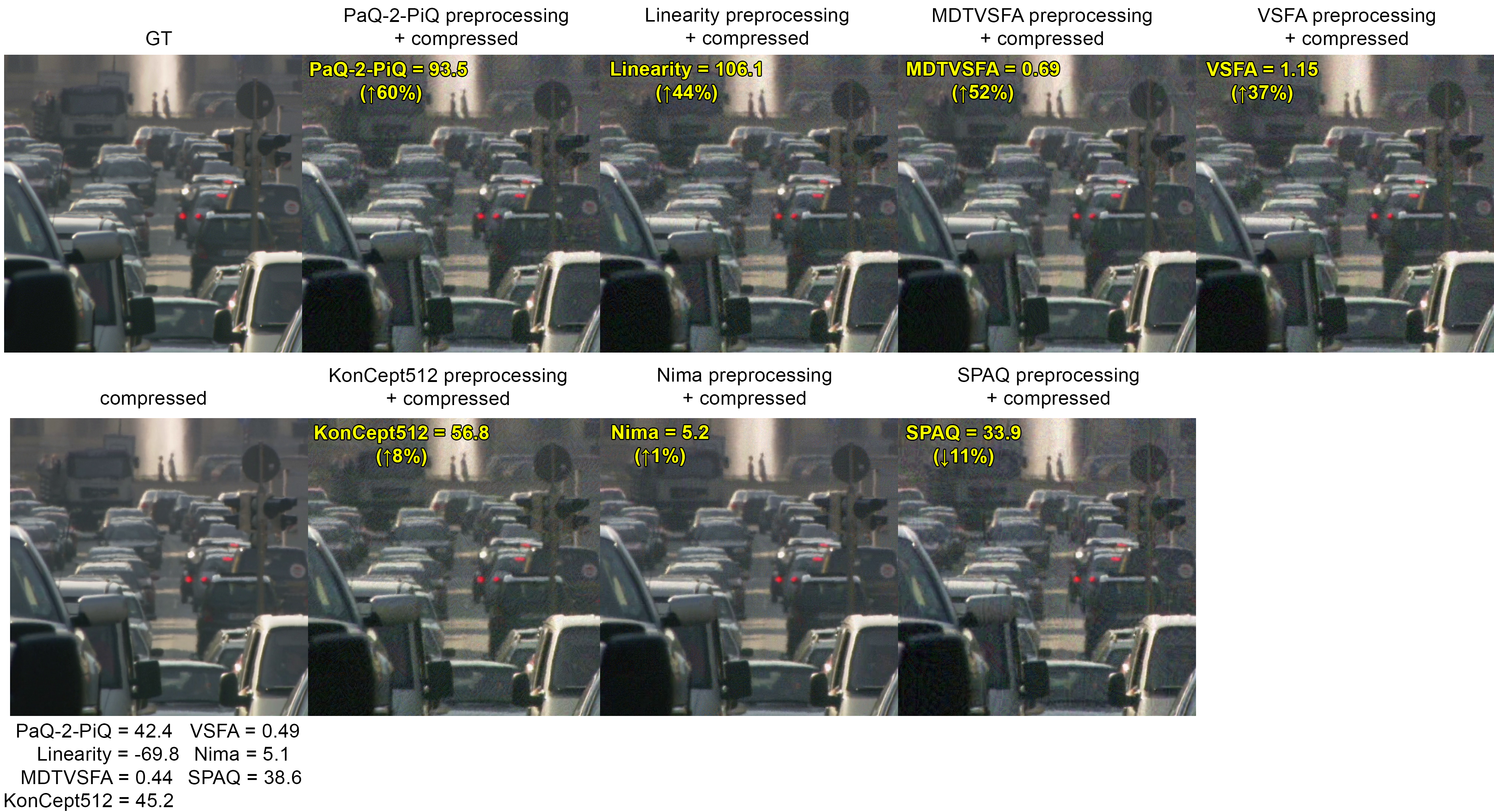}
\end{center}
   \caption{Comparison of proposed attack on NR metrics applied before compression (H.264 5Mbps preset medium) at 0.04 amplitude level.}
\label{fig:crops}
\end{figure*}

For each metric we applied universal perturbation attack and tested its stability using the method from Section \ref{sec:stability_score}. Figure \ref{fig:crops} shows attacked video frames after compression. The first column in Table \ref{table:attacks} presents the metric rating by our proposed stability score. Higher scores correspond to more stable metrics. According to the results, four tested NR metrics (PaQ-2-PiQ, Linearity, VSFA and MDTVSFA) proved vulnerable to universal perturbation attacks, whereas three (KonCept512, Nima and SPAQ) proved stable. The second and third columns contain SRCC correlations from subjective comparisons conducted using the MSU and BVQA benchmarks. The stability score characterizes the metric’s reliability; SRCC correlation characterizes the metric’s accuracy on a given dataset. MDTVSFA is clearly the best when judged by SRCC, but it is vulnerable to universal perturbation attack. 

\begin{table}[htb]
\begin{center}
\begin{tabular}{|c |c | c| c|}
\hline
 & Stability score $\uparrow$ & \multicolumn{2}{c}{Spearman correlation coefficient $\uparrow$} \\
\cline{3-4}
& & MSU benchmark \cite{MSU_benchmark} & BVQA benchmark \cite{Bovik_benchmark} \\
\hline
PaQ-2-PiQ \cite{Ying_2020_CVPR} & -5.3 & 0.89 & 0.61 \\
\hline
Linearity \cite{li2020norm} & -4.2 & 0.92 & - \\
\hline
VSFA \cite{li2019quality} & -3.8 & 0.91 & 0.75 \\
\hline
MDTVSFA \cite{li2021unified} & -4.8 & \textbf{0.94} & \textbf{0.78} \\
\hline
KonCept512 \cite{hosu2020koniq} & -0.3 & 0.86 & 0.73 \\
\hline
SPAQ \cite{Fang_2020_CVPR} & \textbf{2.6} & - & - \\
\hline
Nima \cite{talebi2018nima} & -0.1 & 0.85 & - \\
\hline
\end{tabular}
\caption{Stability scores and SRCC correlations for all tested metrics. Our rating shows which metrics are stable and which are easily increased by attack.}
\label{table:attacks}
\end{center}
\end{table}

\section{Conclusion}
In this paper we showed that NR quality metrics can be vulnerable to universal perturbation that is trained to increase their scores when added to original images or videos. We also used trained universal perturbations as preprocessing before video compression and proposed a method for stability assessment of NR quality metrics by stability score based on RD curves. We calculated the stability score for seven popular NR quality metrics. SPAQ, Nima and KonCept512 proved to be resistant to universal perturbation attacks, while PaQ-2- PiQ, Linearity, VSFA and MDTVSFA proved vulnerable. Compared with ratings by SRCC subjective correlations from MSU and BVQA benchmarks, our metric-rating stability score approach showed that the best metrics from traditional benchmarks had low stability scores in our benchmark. Videos and RD curves with universal perturbation preprocessing, as well as attack examples with images for all seven metrics, are in the supplemental materials. We make our code publicly available at: \url{https://github.com/katiashh/UAP_Attack_on_Quality_Metrics}. 

Future research will include the analysis of different methods for universal perturbation generation. Also we will investigate how NR quality metric architecture affects its vulnerability to universal perturbation attacks in order to bring insights in further design of NR quality metrics to make them resistant to attacks. 
%Successful universal perturbation therefore puts NR quality metrics at a disadvantage. If it can confuse a metric, the scores become unreliable because they are subject to unfair increases through certain types of processing.

\section{Acknowledgements}
This study was supported by Russian Science Foundation under grant 22-21-00478, \url{https://rscf.ru/en/project/22-21-00478/}

%We compared our attack method with a gradient attack, performed using MAD competition methodology. The universal-perturbation attack’s main advantage compared with a gradient attack is that the former requires only one inference step (without knowledge of the metric architecture), whereas the latter requires several backpropagation steps through the metric architecture. 

%The stability score can therefore serve as another test for NR image- and video-quality metrics to complement traditional subjective tests and to characterize the reliability of the scores those metrics produce.

 %Stability scores measure the metric’s reliability, whereas SRCC subjective correlations measure its accuracy on a given dataset. The stability score can therefore serve as another test for NR image- and video-quality metrics to complement traditional subjective tests and to characterize the reliability of the scores those metrics produce.
%It turns out that a metric’s accuracy fails to guarantee stability.

\bibliography{egbib}
\end{document}